\documentclass[runningheads]{llncs}
\usepackage[T1]{fontenc}
\usepackage{xcolor}
\usepackage{graphicx,verbatim}
\usepackage{amsmath}
\usepackage{amssymb}

\usepackage{hyperref}
\usepackage{color}

\begin{document}

\title{Prototype-Based Multiple Instance Learning for Gigapixel Whole Slide Image Classification}
\titlerunning {ProtoMIL for WSI Classification}

%
\author{Susu Sun\inst{1}\and
Dominique van Midden\inst{2} \and 
Geert Litjens\inst{2,3} \and \\
Christian F. Baumgartner\inst{1,4}}
\authorrunning{S. Sun et al.}
%
\institute{Cluster of Excellence: Machine Learning - New Perspectives for Science,\\ University of Tübingen, Germany \\ \email{susu.sun@uni-tuebingen.de}\and
Radboud University Medical Center, Netherlands \\
\email{\{dominique.vanmidden, geert.litjens\}@radboudumc.nl}\and
Oncode Institute, Netherlands \and 
Faculty of Health Sciences and Medicine, University of Lucerne, Switzerland \\
\email{christian.baumgartner@unilu.ch}\\
}



\maketitle              
\begin{abstract}

Multiple Instance Learning (MIL) methods have succeeded remarkably in histopathology whole slide image (WSI) analysis. However, most MIL models only offer attention-based explanations that do not faithfully capture the model's decision mechanism and do not allow human-model interaction. To address these limitations, we introduce ProtoMIL, an inherently interpretable MIL model for WSI analysis that offers user-friendly explanations and supports human intervention\footnote{Code will be publicly available at \url{https://github.com/ss-sun/ProtoMIL}}. Our approach employs a sparse autoencoder to discover human-interpretable concepts from the image feature space, which are then used to train ProtoMIL. The model represents predictions as linear combinations of concepts, making the decision process transparent. Furthermore, ProtoMIL allows users to perform model interventions by altering the input concepts. Experiments on two widely used pathology datasets demonstrate that ProtoMIL achieves a classification performance comparable to state-of-the-art MIL models while offering intuitively understandable explanations. Moreover, we demonstrate that our method can eliminate reliance on diagnostically irrelevant information via human intervention, guiding the model toward being right for the right reason. 


\end{abstract}

\section{Introduction}

\begin{figure}[t]
\includegraphics[width=\textwidth]{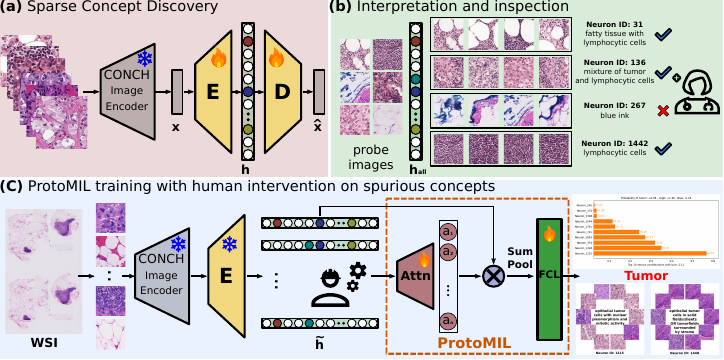}
\caption{The ProtoMIL Pipeline: (a) A sparse autoencoder is trained to discover human-interpretable concepts from the image embedding space. (b) The concepts are linked to visual representations, allowing pathologists to examine them and identify spurious features. (c) ProtoMIL is trained to use the discovered concepts as inputs, with the option to disable spurious concepts via human intervention.}
\label{fig:framework}
\end{figure}

Histopathology whole slide images (WSIs) play an essential role in cancer diagnosis~\cite{vranic2021role}. While deep learning models have the potential to assist pathologists with diagnostic tasks, using them for WSI analysis is challenging due to the large size of such images and the fact that labels are often only available on the slide level. Multiple instance learning (MIL) methods~\cite{ilse2018attention,lu2021data,shao2021transmil,javed2022additive,kapse2024si,sun2025labelfreeconceptbasedmultiple} address these challenges by treating WSI patches as instances and combining patch-level learning with slide-level supervision. 

In safety-critical domains such as computational pathology, interpretability is essential for building trust and effective collaboration between ML algorithms and human users. Traditional MIL models can provide explanations by identifying the most salient regions using intermediate attention scores~\cite{ilse2018attention,lu2021data,shao2021transmil}. However, these spatial explanations provide limited information and suffer from faithfulness issues, as the non-linear relationship between attention scores and the final prediction prevents them from accurately reflecting the model's true decision process~\cite{javed2022additive}.

Inherently interpretable MIL models, such as those proposed in~\cite{javed2022additive,kapse2024si,sun2025labelfreeconceptbasedmultiple}, address the faithfulness problem by enforcing a linear relationship between attention scores and the predictions. Self-interpretable MIL (SI-MIL)~\cite{kapse2024si} further enhances interpretability by transforming the input image into pathology features, such as the statistical properties of nuclei. However, those features are hand-crafted and rely on predictions from auxiliary models. Concept MIL~\cite{sun2025labelfreeconceptbasedmultiple} extends this approach by defining text-based concepts and leveraging vision-language models to predict whether these concepts are shown in the image. Although SI-MIL and Concept MIL offer user-friendly interpretations while maintaining state-of-the-art classification performance, they require predefining pathology concepts, which is often not trivial and limits the flexibility of those approaches. 

In this work, we propose ProtoMIL, an approach to automatically discover pathology concepts and use them to train an inherently interpretable MIL framework. To achieve this, we leverage recent advances in mechanistic interpretability, employing a sparse autoencoder (SAE)~\cite{bricken2023towards,rao2024discover,le2024learning} to learn pathology concepts automatically. These discovered concepts can be inspected through representative patches from a probing set, allowing users to understand the concepts better and identify diagnostically irrelevant ones. ProtoMIL is then trained using the discovered concepts, with the option to eliminate unwanted concepts through human intervention. While both concept discovery and human-model interaction have been explored in computer vision~\cite{koh2020concept,steinmann2023learning,shin2023closer,rao2024discover} and natural language processing~\cite{bricken2023towards,cunningham2023sparse}, they are underexplored in computational pathology, especially for WSI analysis. A related approach is ProtoMixer~\cite{butke2023mixing}, which clusters image embeddings to derive prototypes and applies an MLP-Mixer architecture for cancer subtyping, primarily aiming to reduce WSI size and improve computational efficiency. However, its reliance on cluster centroids may limit interpretability, as the prototypes do not necessarily correspond to semantically meaningful pathology concepts. To our knowledge, our framework is the first in computational pathology to demonstrate the effectiveness of automatic concept discovery and interventions on concepts within an MIL framework.

We evaluated ProtoMIL on two widely used histopathology datasets: Camelyon16 and PANDA. Our results show that the concepts learned by the SAE are highly interpretable and align well with pathology features used in clinical practice. Training ProtoMIL with these concepts achieves classification performance comparable to state-of-the-art MIL models. Furthermore, we demonstrate that our method allows the removal of diagnostically irrelevant features via human intervention, which can reduce spurious correlations and steer the model toward being right for the right reason.

\section{Methods}\label{sec:methods}

Our approach consists of three stages. First (Fig.~\ref{fig:framework}a), we train a sparse autoencoder (SAE) to discover human-interpretable concepts from image embeddings generated by the image encoder from CONCH~\cite{lu2024visual} (see Secs.~\ref{sec:extract_img_embd} \& \ref{sec:extract_visual_concepts}). Secondly (Fig.~\ref{fig:framework}b), we use probing images to obtain visual representations for each activated concept neuron. Optionally, pathologists can inspect and name the found concepts based on the representative patches (see Sec.~\ref{sec:obtain_prototypes}). Thirdly (Fig.~\ref{fig:framework}c), we use the same image encoder and learned SAE encoder from the first stage to extract concepts and use them to train our proposed ProtoMIL model leading to interpretable WSI classification (see Sec.~\ref{sec:build_protoMIL}). Additionally, concepts corresponding to diagnostically irrelevant information can be identified by human users in the second stage and removed from the final model in the third stage by intervening on the concept activation vectors (see Sec.~\ref{sec:spurious_signal_removal}).

\subsection{Extracting Image Embeddings}\label{sec:extract_img_embd}

To train the SAE (Fig.~\ref{fig:framework}a) and ProtoMIL (Fig.~\ref{fig:framework}c), we first extract image embeddings for the patches from WSIs. For this step, we follow the pipeline of CLAM~\cite{lu2021data}, a widely used MIL model. We first segment the tissue regions and background in each WSI, then divide the tissue regions into $N$ non-overlapping patches. We use a patch size of $256 \times 256$ at the resolution level of $0.5 \mu m / \text{pixel}$. Those patches are then fed into a feature extractor to obtain image embeddings with a dimensionality of 512. For this step, we employ the pre-trained and frozen CONCH image encoder~\cite{lu2024visual}, a state-of-the-art histopathology foundation model.

\subsection{Discovering Sparse Human Interpretable Concepts}\label{sec:extract_visual_concepts}

To automatically discover human-interpretable concepts, we train an SAE in which the latent representation has \emph{larger} dimensionality than the input feature (see Fig.~\ref{fig:framework}a). Following~\cite{rao2024discover,bricken2023towards}, our SAE consists of an encoder $E$ and a decoder $D$. The encoder takes as input an image embedding vector $\vec{x} \in \mathbb{R}^{d_\text{in}}$ obtained using CONCH, and produces a latent representation $\vec{h} \in \mathbb{R}^{d_\text{hid}}$ via a linear transformation with weight matrix $\mathbf{W}_{\text{enc}}$ and bias vector $\vec{b}$,
\begin{equation}
\vec{h} = \text{ReLU}(\mathbf{W}_{\text{enc}} \vec{x} + \vec{b}).
\end{equation}
Crucially, here $d_\text{hid} > d_\text{in}$ to allow for a sparse representation. 
The decoder is also a linear operation and reconstructs the input from $\vec{h}$ as 
\begin{equation}
\vec{\hat{x}} = \mathbf{W}_{\text{dec}} \vec{h} = \sum_{i=1}^{d_{\text{hid}}} h_i \vec{f_i},
\end{equation}
where $\mathbf{W}_\text{dec}$ is a weight matrix with row vectors $\vec{f_i}$, which have been shown associated with highly interpretable concepts in~\cite{bricken2023towards}. The reconstruction $\vec{\hat{x}}$ can be decomposed as a linear combination of these concepts, with $h_i$ capturing how much that concept is activated. We refer to $\vec{h}$ as the concept activation vector.

We train the SAE to minimize an $L_2$ reconstruction loss between the input and its reconstruction, along with an $L_1$ sparsity regularization term which encourages the latent vector $\vec{h}$ to be sparse:
\begin{equation}
\mathcal{L}_{\text{SAE}}(\vec{x}) = \| \vec{\hat{x}} - \vec{x} \|_2^2 + \lambda_1 \| \vec{h} \|_1.
\label{eq:sae_loss}
\end{equation}

Although the above training loss does not guarantee that $\vec{f_i}$ learned by SAE correspond to human-interpretable or mono-semantic concepts, prior studies across different domains~\cite{bricken2023towards,cunningham2023sparse,rao2024discover,le2024learning} have shown that the neurons in the latent representation often do align with distinct and interpretable concepts. In preliminary experiments, we searched $\lambda_1$ over \{$1\times10^{-4}$, $3\times10^{-4}$, $5\times10^{-4}$, $1\times10^{-3}$\} and $d_{hid}$ over \{1024, 2048, 4096\} and found that training the SAE with $\lambda_1 = 3\times10^{-4}$ and $d_{hid} = 2048$ gave sufficiently sparse and interpretable concepts, and we used those values for our experiments.



\subsection{Associating Concepts with Prototypical Visual Representations}\label{sec:obtain_prototypes}

\begin{figure}[t!]
\centering
\includegraphics[width=0.95\textwidth]{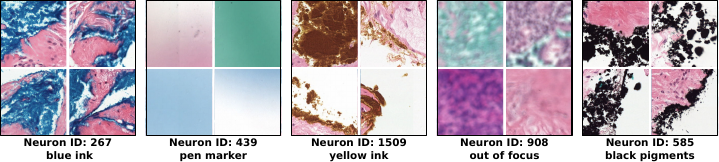}
\caption{Concepts correspond to artifacts in the PANDA dataset identified by our SAE.}
\label{fig:artifacts}
\end{figure}

The concepts $\vec{f_i}$ discovered by the SAE are not yet directly interpretable by humans. To associate the concepts with visual representations, we construct a probing image set by randomly selecting 10,000 normal patches and 10,000 tumor patches from the training set of downstream tasks. We compute the concept activation vector $\vec{h}$ for all patches and identify $\vec{f_i}$ as an activated concept if its corresponding $h_i$ has at least one non-zero value across the entire probing set. 

\begin{figure}[t!]
\centering
\includegraphics[width=0.95\textwidth]{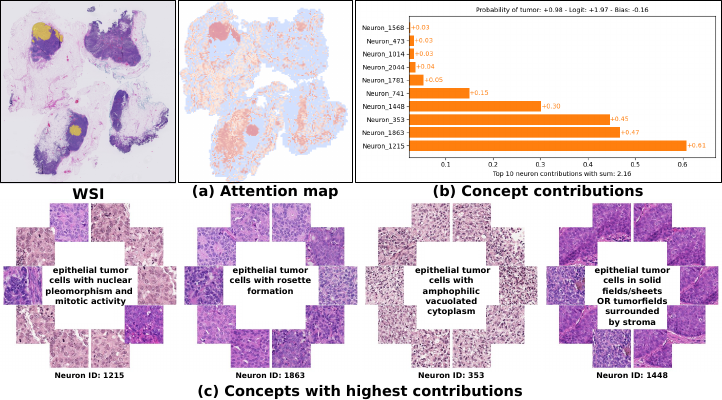}
\caption{Local explanation for a WSI from Camelyon16, with ground truth tumor regions in yellow. The explanation consists of three components: (a) attention map, (b) concept contribution vector, and (c) prototypical representations for key concepts.}
\label{fig:local_exps}
\end{figure}

We then construct prototypical visual representations for the activated concepts by selecting the 10 patches from the probing set with the highest activation values for each concept. Those patches capture the visual patterns characteristic of the respective concept $\vec{f_i}$. We additionally ask pathologists to name the concepts by presenting them with those prototypical image patches. Figs. \ref{fig:artifacts} and \ref{fig:local_exps}c show some examples of prototypical representations and the description of the respective concept by pathologists.

\subsection{Constructing ProtoMIL models}\label{sec:build_protoMIL}

We train our inherently interpretable MIL model using concept activation vectors $\vec{h}$ as input. Given an input WSI, we first extract image embeddings for all patches using the CONCH image encoder. These embeddings are then transformed into concept activation vectors through the frozen encoder of our trained SAE. The resulting vectors are then used to train our ProtoMIL model. 

As illustrated in Fig.~\ref{fig:framework}c, ProtoMIL uses a comparatively simple architecture consisting of the attention module introduced in~\cite{javed2022additive}, a sum pooling function, and a linear classifier. 
It takes the interpretable concept activation vectors $\vec{h_p}$ for each patch $p \in [1, N]$ in the WSI as inputs. The attention module calculates softmax attention scores $a_p$ for each patch, which are then used to weight the corresponding vectors $\vec{h_p}$. A pooling operation is then performed by summing all weighted vectors across specific concepts. Finally, a linear classifier implemented as a single fully connected layer followed by a softmax is used to generate the final prediction. 
The logit output of a target class $c$ can be expressed as
\begin{equation}
    \hat{Y}_{c} = \sum_{i=1}^{d_{hid}} \sum_{p=1}^{N} w_{ci}  a_p h_{pi} + b_c = \sum_{i=1}^{d_{hid}} \kappa_{ci} + b_c,
\end{equation}
where $h_{pi}$ denotes the value of the $i$-th dimension of input concept activation vector $\vec{h_p}$, and the parameters $w_{ci}$ and $b_c$ correspond to the weights and bias of class $c$ in the linear classifier. The term $\kappa_{ci} =  \sum_{p=1}^{N}w_{ci}  a_p h_{pi}$, captures the contributions of concept $i$ to the prediction of class $c$ on the WSI level.

ProtoMIL provides a local explanation for individual WSI predictions, which consists of three components. First, we visualize the attention scores $a_p$ as an attention map (see Fig.~\ref{fig:local_exps}a) to highlight the key regions. This is similar to previous MIL approaches such as~\cite{javed2022additive,kapse2024si,sun2025labelfreeconceptbasedmultiple}. Going one step further, we visualize the contributions $\kappa_{ci}$ of the most influential concepts for this WSI prediction (see Fig.~\ref{fig:local_exps}b). Moreover, the prototypical representations for these key concepts are displayed (see Fig.~\ref{fig:local_exps}c) to allow the operating pathologist to understand how ProtoMIL arrived at a prediction. 

Additionally, ProtoMIL offers a global explanation for its decision mechanism at the dataset level. This is achieved by computing each class's mean concept contributions vectors over entire datasets (Fig.~\ref{fig:global_exp}). This approach allows the users to assess the model's overall behavior and quality.

\subsection{Removing Spurious Signals}\label{sec:spurious_signal_removal}

Neural networks can rely on spurious signals and task-irrelevant features for making predictions~\cite{geirhos2020shortcut,adebayo2022post,sun2024attri,sun2023right}. WSIs often contain artifacts such as inks, air bubbles, or out-of-focus regions that may introduce spurious correlations to the neural networks. As we will show in Sec.\,\ref{sec:experiments}, the SAE does detect artifacts in the data we use. Our proposed ProtoMIL approach provides a convenient way to disable dependence on such unwanted spurious signals. We achieve this by identifying the concept neurons corresponding to the spurious signals (Fig.~\ref{fig:framework}b), and then explicitly setting their activation values $h_i$ to zero during the ProtoMIL training (Fig.~\ref{fig:framework}c). As we will show in Sec.\,\ref{sec:exp_eval_spurious_removal_train}, this allows the model to be trained using only desirable class-relevant concepts and being right for the right reason.

\section{Experiments and Results}
\label{sec:experiments}

\subsubsection{Data and Baselines} We evaluated our approach on two widely used histopathology WSI datasets: Camelyon16~\cite{bejnordi2017diagnostic} for breast cancer metastasis detection and PANDA for prostate cancer grading~\cite{Bulten2022}. We used the official test set of Camelyon16 for evaluation and split the 270 WSIs in the official training set into training and validation sets with an 80/20 ratio. Since the official PANDA test set is not publicly available, we split the 10616 WSIs in the official training set into training, validation, and test sets with an 80/10/10 ratio. 

We compared our method to four state-of-the-art MIL models:  Attention MIL~\cite{ilse2018attention}, CLAM~\cite{lu2021data}, TransMIL~\cite{shao2021transmil} and Additive MIL~\cite{javed2022additive}. 

\subsubsection{Implementation and Training}

We trained our SAE model with a learning rate of 0.0001 for 60 epochs. To identify generalized concepts applicable to both datasets, we trained the SAE on data from both Camelyon16 and PANDA. We trained our ProtoMIL and baseline models for both datasets for binary classification using a learning rate of $1\times10^{-4}$, cross-entropy loss, and 200 epochs. We used the validation sets for model selection.

\subsubsection{Evaluations of Concepts Discovery}\label{sec:exp_eval_visual_concepts}

We found that the 2048 dimensional concept activation vectors $\vec{h}$ are highly sparse, with an average of 10 non-zero values per vector. Across the entire SAE training set, which contains 3 million patches from Camelyon16 and 4.7 million patches from PANDA, a total of 74 concepts were activated.  
We analyzed the discovered concepts (Fig.~\ref{fig:artifacts}, Fig.~\ref{fig:local_exps}c) in collaboration with a pathologist. By examining the prototypical representations, we confirmed that the SAE effectively identified concepts linked to the pathology features, such as glandular structure, cellular characteristics, and artifacts, while being robust to variations in staining methods. Although Camelyon16 and PANDA target different diseases in different organs, we found that certain neurons can capture shared pathology features across both datasets. According to the pathologist's evaluation, 75\% of the detected concepts on Camelyon16 and 86\% on PANDA captured monosemantic pathology features.

\begin{figure}[t!]
\includegraphics[width=\textwidth]{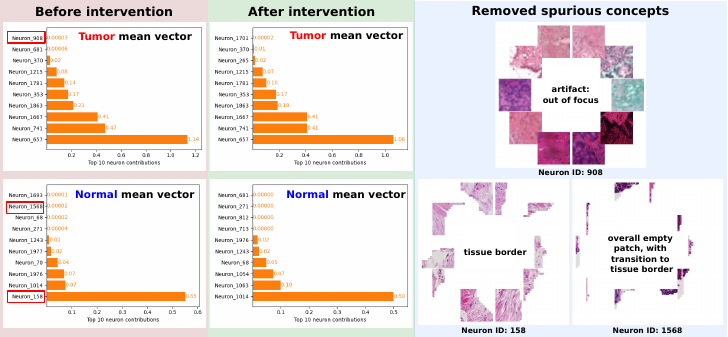}
\caption{Global explanations for ProtoMIL models on the PANDA dataset before and after removing spurious features. The mean vectors on the left show the top 10 concept neurons' average contributions to the tumor and normal classes. Spurious concepts excluded after human intervention are highlighted on the right.}
\label{fig:global_exp}
\end{figure}

\begin{table}[t!]
    \caption{Classification performance measured by Accuracy and AUC. }
    \label{tab:classification}
    \centering
    \begin{tabular}{c|cc|cc}
        \hline
         \textbf{Model} & \multicolumn{2}{c|}{\textbf{Camelyon16}} & \multicolumn{2}{c}{\textbf{PANDA}} \\
         & Acc. & AUC & Acc. & AUC \\
        \hline
        ABMIL (image) &0.922 &0.908  &0.892  &0.953  \\
        CLAM (image) &0.915 &0.966  &0.884  &0.979  \\
        TransMIL (image) &0.938  &0.950  &0.939  &0.977 \\
        AdditiveMIL (image) & 0.875 & 0.883 &0.905 &0.958\\
        ProtoMIL (concept) &0.907 &0.918  &0.916  &0.970  \\
        ProtoMIL (intervened concepts) &0.926 &0.913   &0.916  &0.964  \\
        \hline
    \end{tabular}
\end{table}

\subsubsection{Evaluation without Interventions}\label{sec:exp_eval_normal_train}

We first trained ProtoMIL using the concept activation vectors $\vec{h}$ as input without interventions. We present the classification performance of ProtoMIL and baseline models in Table~\ref{tab:classification}. The results show that ProtoMIL 
achieved classification performance comparable to conventional MIL models trained on image features.

Figure~\ref{fig:local_exps} shows ProtoMIL's local explanation for an input WSI. We found that regions with high attention scores in Fig.~\ref{fig:local_exps}a coincide with the ground truth tumor regions. Moreover, we verified that the key concepts are all related to tumor features using the prototypical concept representations (see Fig.~\ref{fig:local_exps}c).

\subsubsection{Evaluations with Intervention on Spurious Concepts}\label{sec:exp_eval_spurious_removal_train}

We examined all 74 concepts learned by SAE and identified five concepts associated with artifacts in both datasets (see Fig.~\ref{fig:artifacts}). By generating the global explanations for the ProtoMIL trained without intervention (see Fig.~\ref{fig:global_exp}), we found that three concepts among the top 10 contributors do not contain diagnostic information. These concepts were associated with out-of-focus artifacts and tissue borders. 

We set the activation values of these detected artifacts and diagnostically irrelevant concepts to zero as described in Sec.\,\ref{sec:spurious_signal_removal} and retrained ProtoMIL to remove the spurious correlation. As seen in Table~\ref{tab:classification}, ProtoMIL maintains a strong classification performance despite the intervention. In the global explanations of the model trained after the intervention (Fig.~\ref{fig:global_exp}), it can be seen that -- as expected -- the spurious concepts no longer affect the predictions. We confirmed with a pathologist that the concepts used by ProtoMIL after the intervention are more class-relevant, indicating improved model quality.

\section{Discussion and Conclusion}
This paper proposes a novel, inherently interpretable MIL model for WSI analysis. Our experimental results show that SAE effectively learns sparse, human-interpretable pathology concepts. Using the activations of the learned concepts, we train an inherently interpretable ProtoMIL model, which achieves comparable classification performance with state-of-the-art MIL models while providing easy-to-understand explanations. A key advantage of our approach is that it allows model interventions to eliminate dependence on spurious signals, potentially improving the model's reliability. As we have observed that a certain number of neurons in the SAE remain polysemantic, in future work, we plan to conduct a more thorough investigation of the learned concept space and develop methods to further improve the SAE and reduce polysemantic representations. Additionally, we plan to explore the feasibility of developing a more generalized SAE capable of learning transferable concepts across diverse tissue categories and cancer types, thereby providing interpretable concept representations for a wide range of downstream tasks.


\begin{credits}

\subsubsection{\ackname}
Funded by the Deutsche Forschungsgemeinschaft (DFG) – EXC number 2064/1 – Project number 390727645, the Carl Zeiss Foundation in the project ``Certification and Foundations of Safe Machine Learning Systems in Healthcare'', the Dutch Research Council, the European Research Council, the European Union Horizon Europe Program,
and the Hanarth Fund. The authors thank the International Max Planck Research School for Intelligent Systems (IMPRS-IS) for supporting Susu Sun.

\end{credits}

\newpage
\bibliographystyle{splncs04}
\bibliography{mybibliography}

\begin{thebibliography}{10}
\providecommand{\url}[1]{\texttt{#1}}
\providecommand{\urlprefix}{URL }
\providecommand{\doi}[1]{https://doi.org/#1}

\bibitem{adebayo2022post}
Adebayo, J., Muelly, M., Abelson, H., Kim, B.: Post hoc explanations may be ineffective for detecting unknown spurious correlation. In: International conference on learning representations (2022)

\bibitem{bejnordi2017diagnostic}
Bejnordi, B.E., Veta, M., Van~Diest, P.J., Van~Ginneken, B., Karssemeijer, N., Litjens, G., Van Der~Laak, J.A., Hermsen, M., Manson, Q.F., Balkenhol, M., et~al.: Diagnostic assessment of deep learning algorithms for detection of lymph node metastases in women with breast cancer. Jama  \textbf{318}(22),  2199--2210 (2017)

\bibitem{bricken2023towards}
Bricken, T., Templeton, A., Batson, J., Chen, B., Jermyn, A., Conerly, T., Turner, N., Anil, C., Denison, C., Askell, A., et~al.: Towards monosemanticity: Decomposing language models with dictionary learning. Transformer Circuits Thread  \textbf{2} (2023)

\bibitem{Bulten2022}
Bulten, W., Kartasalo, K., Chen, P.H.C., Str\"{o}m, P., Pinckaers, H., Nagpal, K., Cai, Y., Steiner, D.F., van Boven, H., Vink, R., van~de Kaa, C.H., van~der Laak, J., Amin, M.B., Evans, A.J., van~der Kwast, T., Allan, R., Humphrey, P.A., Gr\"{o}nberg, H., Samaratunga, H., Delahunt, B., Tsuzuki, T., H\"{a}kkinen, T., Egevad, L., Demkin, M., Dane, S., Tan, F., Valkonen, M., Corrado, G.S., Peng, L., Mermel, C.H., Ruusuvuori, P., Litjens, G., Eklund, M., {The PANDA challenge consortium}: Artificial intelligence for diagnosis and gleason grading of prostate cancer: the {PANDA} challenge. Nature Medicine  (Jan 2022). \doi{10.1038/s41591-021-01620-2}, \url{https://doi.org/10.1038/s41591-021-01620-2}

\bibitem{butke2023mixing}
Butke, J., Hashimoto, N., Takeuchi, I., Miyoshi, H., Ohshima, K., Sakuma, J.: Mixing histopathology prototypes into robust slide-level representations for cancer subtyping. In: International Workshop on Machine Learning in Medical Imaging. pp. 114--123. Springer (2023)

\bibitem{cunningham2023sparse}
Cunningham, H., Ewart, A., Riggs, L., Huben, R., Sharkey, L.: Sparse autoencoders find highly interpretable features in language models. arXiv preprint arXiv:2309.08600  (2023)

\bibitem{geirhos2020shortcut}
Geirhos, R., Jacobsen, J.H., Michaelis, C., Zemel, R., Brendel, W., Bethge, M., Wichmann, F.A.: Shortcut learning in deep neural networks. Nature Machine Intelligence  \textbf{2}(11),  665--673 (2020)

\bibitem{ilse2018attention}
Ilse, M., Tomczak, J., Welling, M.: Attention-based deep multiple instance learning. In: International conference on machine learning. pp. 2127--2136. PMLR (2018)

\bibitem{javed2022additive}
Javed, S.A., Juyal, D., Padigela, H., Taylor-Weiner, A., Yu, L., Prakash, A.: Additive {MIL}: Intrinsically interpretable multiple instance learning for pathology. Advances in Neural Information Processing Systems  \textbf{35},  20689--20702 (2022)

\bibitem{kapse2024si}
Kapse, S., Pati, P., Das, S., Zhang, J., Chen, C., Vakalopoulou, M., Saltz, J., Samaras, D., Gupta, R.R., Prasanna, P.: {SI-MIL}: Taming deep {MIL} for self-interpretability in gigapixel histopathology. In: Proceedings of the IEEE/CVF Conference on Computer Vision and Pattern Recognition. pp. 11226--11237 (2024)

\bibitem{koh2020concept}
Koh, P.W., Nguyen, T., Tang, Y.S., Mussmann, S., Pierson, E., Kim, B., Liang, P.: Concept bottleneck models. In: International conference on machine learning. pp. 5338--5348. PMLR (2020)

\bibitem{le2024learning}
Le, N.M., Patel, N., Shen, C., Martin, B., Eng, A., Shah, C., Grullon, S., Juyal, D.: Learning biologically relevant features in a pathology foundation model using sparse autoencoders. In: Advancements In Medical Foundation Models: Explainability, Robustness, Security, and Beyond (2024), \url{https://openreview.net/forum?id=daV16mhUBd}

\bibitem{lu2024visual}
Lu, M.Y., Chen, B., Williamson, D.F., Chen, R.J., Liang, I., Ding, T., Jaume, G., Odintsov, I., Le, L.P., Gerber, G., et~al.: A visual-language foundation model for computational pathology. Nature Medicine  \textbf{30}(3),  863--874 (2024)

\bibitem{lu2021data}
Lu, M.Y., Williamson, D.F., Chen, T.Y., Chen, R.J., Barbieri, M., Mahmood, F.: Data-efficient and weakly supervised computational pathology on whole-slide images. Nature biomedical engineering  \textbf{5}(6),  555--570 (2021)

\bibitem{rao2024discover}
Rao, S., Mahajan, S., B{\"o}hle, M., Schiele, B.: Discover-then-name: Task-agnostic concept bottlenecks via automated concept discovery. In: European Conference on Computer Vision. pp. 444--461. Springer (2024)

\bibitem{shao2021transmil}
Shao, Z., Bian, H., Chen, Y., Wang, Y., Zhang, J., Ji, X., et~al.: Trans{MIL}: Transformer based correlated multiple instance learning for whole slide image classification. Advances in neural information processing systems  \textbf{34},  2136--2147 (2021)

\bibitem{shin2023closer}
Shin, S., Jo, Y., Ahn, S., Lee, N.: A closer look at the intervention procedure of concept bottleneck models. In: International Conference on Machine Learning. pp. 31504--31520. PMLR (2023)

\bibitem{steinmann2023learning}
Steinmann, D., Stammer, W., Friedrich, F., Kersting, K.: Learning to intervene on concept bottlenecks. arXiv preprint arXiv:2308.13453  (2023)

\bibitem{sun2023right}
Sun, S., Koch, L.M., Baumgartner, C.F.: Right for the wrong reason: Can interpretable ml techniques detect spurious correlations? In: International Conference on Medical Image Computing and Computer-Assisted Intervention. pp. 425--434. Springer (2023)

\bibitem{sun2025labelfreeconceptbasedmultiple}
Sun, S., Tessier, L., Meeuwsen, F., Grisi, C., van Midden, D., Litjens, G., Baumgartner, C.F.: Label-free concept based multiple instance learning for gigapixel histopathology (2025), \url{https://arxiv.org/abs/2501.02922}

\bibitem{sun2024attri}
Sun, S., Woerner, S., Maier, A., Koch, L.M., Baumgartner, C.F.: Attri-net: A globally and locally inherently interpretable model for multi-label classification using class-specific counterfactuals. arXiv preprint arXiv:2406.05477  (2024)

\bibitem{vranic2021role}
Vranic, S., Gatalica, Z.: The role of pathology in the era of personalized (precision) medicine: A brief review.  (2021)

\end{thebibliography}

\end{document}